\documentclass[11pt,a4paper]{article}

\usepackage[utf8]{inputenc}
\usepackage[T1]{fontenc}
\usepackage[english]{babel}
\usepackage{amsmath,amssymb}
\usepackage{graphicx}
\usepackage{booktabs}
\usepackage{tabularx}
\usepackage{hyperref}
\usepackage{cleveref}
\usepackage{listings}
\usepackage{xcolor}
\usepackage{geometry}
\usepackage{enumitem}
\usepackage{float}
\usepackage{caption}
\usepackage{subcaption}
\usepackage{tikz}
\usetikzlibrary{arrows.meta, positioning, shapes.geometric, fit, calc}

\hypersetup{
    colorlinks=true,
    linkcolor=blue!60!black,
    citecolor=blue!60!black,
    urlcolor=blue!60!black
}

\crefname{figure}{Figure}{Figures}
\Crefname{figure}{Figure}{Figures}
\crefname{table}{Table}{Tables}
\Crefname{table}{Table}{Tables}
\crefname{section}{Section}{Sections}
\Crefname{section}{Section}{Sections}
\crefname{appendix}{Appendix}{Appendices}
\Crefname{appendix}{Appendix}{Appendices}

\title{\textbf{ESAA: Event Sourcing for Autonomous Agents in LLM-Based Software Engineering}}
\author{
    Elzo Brito dos Santos Filho \\
    \texttt{elzo.santos@cps.sp.gov.br}
}
\date{February 2026}

\begin{document}

\maketitle

\begin{abstract}
Autonomous agents based on Large Language Models (LLMs) have evolved from reactive assistants to systems capable of planning, executing actions via tools, and iterating over environment observations. However, they remain vulnerable to structural limitations: lack of native state, context degradation over long horizons, and the gap between probabilistic generation and deterministic execution requirements~\cite{yao2022react, liu2024lost}. This paper presents the ESAA (Event Sourcing for Autonomous Agents) architecture, which separates the agent's cognitive intention from the project's state mutation, inspired by the Event Sourcing pattern~\cite{fowler2005event}. In ESAA, agents emit only structured intentions in validated JSON (\texttt{agent.result} or \texttt{issue.report}); a deterministic orchestrator validates, persists events in an append-only log (\texttt{activity.jsonl}), applies file-writing effects, and projects a verifiable materialized view (\texttt{roadmap.json}). The proposal incorporates boundary contracts (\texttt{AGENT\_CONTRACT.yaml}), metaprompting profiles (PARCER), and replay verification with hashing (\texttt{esaa verify}), ensuring the immutability of completed tasks and forensic traceability. Two case studies validate the architecture: (i)~a landing page project (9~tasks, 49~events, single-agent composition) and (ii)~a clinical dashboard system (50~tasks, 86~events, 4~concurrent agents across 8~phases), both concluding with \texttt{run.status=success} and \texttt{verify\_status=ok}. The multi-agent case study demonstrates real concurrent orchestration with heterogeneous LLMs (Claude Sonnet~4.6, Codex GPT-5, Antigravity/Gemini~3~Pro, and Claude Opus~4.6), providing empirical evidence of the architecture's scalability beyond single-agent scenarios.

\end{abstract}

\section{Introduction}
\label{sec:intro}

LLM-based software engineering is shifting from conversational interactions to agentic workflows that require \emph{long-horizon consistency}—that is, maintaining coherence between goals, contracts, and effects over dozens of cycles and artifacts~\cite{yao2022react}. In brownfield scenarios, where the agent needs to navigate large repositories and modify multiple files with cross-dependencies, the risk of \emph{state drift} increases: the agent may believe it has fixed a bug while the actual system remains unchanged, or it may rewrite specifications to bypass local compilation failures.

A recurring empirical source for these difficulties is software engineering benchmarks on real GitHub issues, such as SWE-bench, which aims to measure the ability of agents to produce coherent and verifiable patches in real-world codebases, moving beyond synthetic and isolated problems~\cite{jimenez2024swe}. Even when there is progress in configurations with human validation or restricted suites, the operational challenge of governance remains: auditing what the agent did, why it did it, and how to revert with forensic precision~\cite{openai2024swe}.

Contemporary multi-agent frameworks like AutoGen~\cite{wu2023autogen}, MetaGPT~\cite{hong2023metagpt}, and LangGraph~\cite{langgraph2024} have made significant progress in agent coordination, role assignment, and workflow orchestration. However, these systems typically manage state through in-memory data structures or database snapshots, lacking the \emph{immutable audit trail} and \emph{deterministic replay} guarantees that are essential for production software engineering workflows where accountability and reversibility are non-negotiable.

This work proposes that the core problem is not just "improving the prompt"; it is restructuring the system around verifiable invariants. ESAA applies the Event Sourcing pattern to the agent's lifecycle: the source of truth is not the current snapshot of the repository, but an immutable log of intentions, decisions, and effects, from which the current state is deterministically projected~\cite{fowler2005event}. In parallel, ESAA adopts Command Query Responsibility Segregation principles, aligned with the CQRS tradition, to reduce coupling between writing (changes) and reading (derived state)~\cite{fowler2011cqrs}.
The architecture is validated through two case studies of increasing complexity: a landing page project orchestrated with single-agent composition (9~tasks, 49~events) and a clinical dashboard system (\texttt{clinic-asr}) orchestrated with four heterogeneous LLM agents operating concurrently (50~tasks, 86~events, 8~completed phases). Both cases demonstrate state reproducibility via replay and hash verification.

\section{Related Work}
\label{sec:related}

\subsection{Tool-Using Agents and Interleaved Reasoning}

The ReAct pattern describes a cycle where the model alternates between reasoning and acting via tools, incorporating environment observations into the next step~\cite{yao2022react}. While this interaction increases capability in tool-oriented tasks, it also amplifies vulnerabilities: noisy observations and tool outputs consume the context window, and errors can propagate through multiple turns.

Benchmarks like SWE-bench formalize part of this problem by requiring patches that pass tests in real repositories, capturing a relevant fraction of real-world requirements~\cite{jimenez2024swe}. "Verified" sets and more rigorous evaluation protocols are also emerging, aimed at reducing false positives from solutions that do not address the root of the defect~\cite{openai2024swe}. Nevertheless, these benchmarks do not resolve the need for a governance kernel to control effects and maintain an audit trail.

\subsection{Multi-Agent Frameworks}

Recent multi-agent systems have explored various coordination strategies. AutoGen~\cite{wu2023autogen} enables conversational patterns between agents with role-based specialization. MetaGPT~\cite{hong2023metagpt} assigns software engineering roles (architect, developer, tester) to different agents with structured output protocols. LangGraph~\cite{langgraph2024} provides a graph-based state machine to define agent workflows with conditional edges and cycles.

Although these frameworks advance agent coordination, they share common limitations from an auditability perspective: (i)~state is typically mutable and stored as snapshots rather than a change log; (ii)~there is no native mechanism for deterministic replay of the complete decision trail; (iii)~the blast radius of a compromised agent is not formally bounded by contracts. ESAA addresses these gaps by applying event sourcing as the fundamental state management paradigm. \Cref{tab:framework_comparison} presents a structured comparison.

\begin{table}[ht]
\centering
\caption{Comparison of ESAA with contemporary multi-agent frameworks.}
\label{tab:framework_comparison}
\small
\begin{tabularx}{\textwidth}{lccccc}
\toprule
\textbf{Capability} & \textbf{AutoGen} & \textbf{MetaGPT} & \textbf{LangGraph} & \textbf{CrewAI} & \textbf{ESAA} \\
\midrule
Immutable event log           & --  & --  & --  & --  & \checkmark \\
Deterministic replay          & --  & --  & --  & --  & \checkmark \\
Boundary contracts            & --  & Partial & --  & --  & \checkmark \\
Constrained JSON output       & --  & \checkmark & --  & --  & \checkmark \\
Multi-agent concurrency       & \checkmark & \checkmark & \checkmark & \checkmark & \checkmark \\
Hash-verified projection      & --  & --  & --  & --  & \checkmark \\
Role-based specialization     & \checkmark & \checkmark & \checkmark & \checkmark & \checkmark \\
"Done" immutability rule      & --  & --  & --  & --  & \checkmark \\
Blast radius containment      & --  & Partial & --  & --  & \checkmark \\
\bottomrule
\end{tabularx}
\end{table}

\subsection{Context Limits and Long-Horizon Degradation}

Recent works point to systematic degradation in long context windows, with a drop in recall in the middle of the context (\emph{lost-in-the-middle}), affecting tasks that depend on facts buried in logs, docs, or extensive histories~\cite{liu2024lost}. This limitation creates recency bias and can induce the agent to ignore initial contracts, especially in environments requiring consistency with specifications and security policies.

\subsection{Event Sourcing and CQRS}

Event Sourcing records every state change as an immutable event in an append-only log; the current state is a projection of these events~\cite{fowler2005event}. Alongside this, CQRS separates write and read models to optimize consistency and queries, reducing complexity in systems with strong audit and traceability requirements~\cite{fowler2011cqrs}. ESAA transposes these principles to LLM-assisted software engineering: the agent does not "write state"; it emits intentions and proposed diffs that are validated and applied by a deterministic orchestrator.

\subsection{Constrained Output and Syntactic Structure}

A practical bottleneck in agents is the \emph{structure gap}: models generate probabilistic text, but systems require payloads compliant with schemas (JSON, API calls, patches). Constrained decoding and structured generation approaches guide sampling using grammars and schemas, reducing parsing failures and rejections~\cite{outlines,xgrammar}. Inference and serving frameworks like SGLang also highlight the relevance of this layer by integrating mechanisms for structured outputs and high-throughput execution pipelines~\cite{sglang}.

\section{ESAA Architecture}
\label{sec:architecture}

ESAA establishes a strict separation between (i)~the LLM's heuristic cognition and (ii)~the system's deterministic execution. The agent does not have direct write permission to the project or the event store. Its role is to emit structured intentions and change proposals, always compliant with an output contract validatable by JSON Schema~\cite{jsonschema2020}.

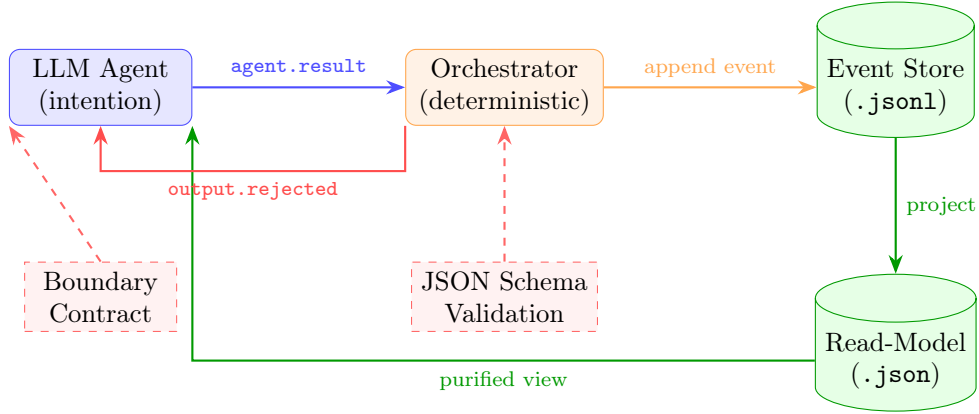
\begin{figure}[ht]
\centering
\begin{tikzpicture}[
    node distance=1.4cm and 2.8cm,
    every node/.style={font=\small},
    agent/.style={rectangle, draw=blue!70, fill=blue!10, rounded corners, minimum width=2.4cm, minimum height=0.9cm, align=center},
    orch/.style={rectangle, draw=orange!70, fill=orange!10, rounded corners, minimum width=2.6cm, minimum height=0.9cm, align=center},
    store/.style={cylinder, draw=green!60!black, fill=green!10, shape border rotate=90, minimum width=2.0cm, minimum height=1.0cm, aspect=0.3, align=center},
    contract/.style={rectangle, draw=red!60, fill=red!5, dashed, minimum width=2.0cm, minimum height=0.7cm, align=center},
    arrow/.style={-{Stealth[length=2.5mm]}, thick}
]
    \node[agent] (agent) {LLM Agent\\(intention)};
    \node[orch, right=2.8cm of agent] (orch) {Orchestrator\\(deterministic)};
    \node[store, right=2.8cm of orch] (eventstore) {Event Store\\(\texttt{.jsonl})};

    \node[contract, below=1.8cm of agent] (contract) {Boundary\\Contract};
    \node[contract, below=1.8cm of orch] (schema) {JSON Schema\\Validation};
    \node[store, below=1.8cm of eventstore] (readmodel) {Read-Model\\(\texttt{.json})};

    \draw[arrow, blue!70] (agent.east) -- node[above, font=\scriptsize] {\texttt{agent.result}} (orch.west);
    \draw[arrow, orange!70] (orch.east) -- node[above, font=\scriptsize] {append event} (eventstore.west);
    \draw[arrow, green!60!black] (eventstore.south) -- node[right, font=\scriptsize] {project} (readmodel.north);
    \draw[arrow, green!60!black] (readmodel.west) -- node[below, font=\scriptsize] {purified view} (agent.south east |- readmodel.west) -- (agent.south east);
    \draw[arrow, red!70] (orch.south west) -- ++(0,-0.6) -| node[pos=0.25, below, font=\scriptsize] {\texttt{output.rejected}} (agent.south);
    \draw[arrow, red!60, dashed] (contract.north) -- (agent.south west);
    \draw[arrow, red!60, dashed] (schema.north) -- (orch.south);
\end{tikzpicture}
\caption{ESAA Orchestration Cycle: the agent emits structured intentions validated by boundary contracts and JSON Schema; the orchestrator persists events in the append-only store and projects the read-model. The agent receives only a purified view, never the raw state.}
\label{fig:architecture}
\end{figure}

\Cref{fig:architecture} illustrates the complete orchestration cycle.

\subsection{Canonical Artifacts}

The ESAA implementation operates with a set of canonical artifacts (stored in \texttt{.roadmap/}):

\begin{enumerate}[label=(\arabic*)]
    \item \textbf{Event store} (\texttt{activity.jsonl}): an append-only log of ordered events (\texttt{event\_seq}), containing intentions, dispatches, effects, and run closures.
    \item \textbf{Materialized view} (\texttt{roadmap.json}): a read-model derived by pure projection; includes tasks, dependencies, indexes, and verification metadata (e.g., \texttt{projection\_hash\_sha256}).
    \item \textbf{Boundary contracts} (\texttt{AGENT\_CONTRACT.yaml}, \texttt{ORCHESTRATOR\_CONTRACT.yaml}): define permitted actions per task type (\texttt{spec}/\texttt{impl}/\texttt{qa}), output patterns, and hard prohibitions (e.g., denying direct \texttt{file.write} by the agent).
    \item \textbf{PARCER profiles} (\texttt{PARCER\_PROFILE.*.yaml}): metaprompting profiles that suppress free-form output and force a strict JSON envelope. PARCER (Persona, Audience, Rules, Context, Execution, Response) defines six dimensions that constrain the agent's output behavior, ensuring each agent receives instructions appropriate to its role with mandatory output schemas and specific prohibited actions per task type.
\end{enumerate}

\subsection{Trace-First Model and Immutability}

ESAA adopts a "trace-first" model: the event is recorded as a fact before any irreversible effect, allowing for audit and containment controls. The \emph{immutability of done rule} defines that completed tasks cannot regress. In case of a defect, an \texttt{issue.report} event creates a new correction path (hotfix) without reopening the history, preserving the decision trail.

\subsection{Determinism through Canonicalization and Hashing}

To ensure reproducibility via replay, ESAA specifies deterministic canonicalization of the projected state and hash calculation. The idea follows the principle of invariant representation for cryptographic operations, as per JSON canonicalization schemes~\cite{rfc8785}. The SHA-256 hash of the canonicalized read-model (\texttt{projection\_hash\_sha256}) allows for the detection of divergence between projection and materialization~\cite{nist2015shs}.

\subsection{Multi-Agent Dispatch and Concurrency}

ESAA supports heterogeneous multi-agent orchestration through the append-only semantics of the event store. The orchestrator assigns tasks to agents based on role-based specialization (e.g., specification agents, implementation agents, QA agents) and tracks claims and completions per agent via correlation identifiers. Concurrent execution is naturally serialized at the event level: multiple agents can work in parallel, but their results are validated and appended sequentially, preserving total ordering in the log. This design allows the orchestrator to detect conflicts (e.g., overlapping file modifications) before applying effects.

\section{Methodology}
\label{sec:methodology}

\subsection{Architectural Intervention Design}

The research follows an artifact-oriented technical validation design: an architecture is proposed (ESAA), formal contracts are defined (schemas, boundaries, event vocabulary), and reproducible execution is demonstrated in two case studies of increasing complexity. The unit of analysis is the orchestrator run, composed of a sequence of events in \texttt{activity.jsonl} and a derivable read-model in \texttt{roadmap.json}.

\subsection{Orchestrator Implementation and Validation Pipeline}

The orchestrator executes a transactional pipeline: (i)~validates agent output via JSON Schema and boundary rules; (ii)~emits \texttt{output.rejected} for violations; (iii)~applies effects via \texttt{orchestrator.file.write}; (iv)~appends events to the event store; (v)~reprojects \texttt{roadmap.json}; (vi)~executes \texttt{esaa verify} via replay and hash comparison. The choice of JSON Schema as a formal contract aims to reduce ambiguity and create observable, treatable failures~\cite{jsonschema2020}.

\subsection{Experimental Setup and Reproducibility}

For open reproducibility purposes, the project's public repository provides the architecture in its initial state (\emph{clean state}), containing only the initialization event (\texttt{run.init}) in the event store. This allows other researchers to execute the pipeline from scratch and observe the deterministic state derivation documented in this study.

\section{Case Studies}
\label{sec:case_studies}

Two case studies validate the ESAA architecture across different complexity scales and agent compositions.

\subsection{Case Study 1: Landing Page (Single-Agent Composition)}
\label{sec:cs1}

The \texttt{example-landingpage} project was structured as a pipeline with 9~sequential tasks (T-1000 to T-1210), covering specification, implementation, and QA. To simulate orchestrator behavior and validate the agents' ability to adhere to constrained output contracts, the execution and artifact generation were conducted using a composition of advanced tools: GPT-5.3-Codex, Claude Code opus~4.6 (integrated with VSCode), and Antigravity (operating with the Gemini~3~Pro model).

The consolidation of this workflow generated a local event store with 49~events, concluding with \texttt{run.status=success} and \texttt{verify\_status=ok}. The produced artifact set included four specifications in \texttt{.roadmap/specs/}, code files in \texttt{src/} (vanilla HTML/CSS/JS), and QA reports in \texttt{.roadmap/qa/}.

The execution followed the cycle: \texttt{attempt.create} $\to$ \texttt{orchestrator.dispatch} $\to$ \texttt{agent.result} $\to$ \texttt{orchestrator.file.write} $\to$ \texttt{task.update} $\to$ (repeat) $\to$ \texttt{verify.ok} $\to$ \texttt{run.end}. This trail materializes the idea that the project state is derived from facts, not direct edits~\cite{fowler2005event}.

\subsection{Case Study 2: Clinical Dashboard (Multi-Agent Concurrency)}
\label{sec:cs2}

The \texttt{clinic-asr} project represents a substantially more complex validation scenario: a clinical dashboard system (POC) with real-time audio transcription capabilities, covering database schema design, REST API contracts, SSE event vocabulary, SPA architecture, configuration management, security policies, and end-to-end testing.

\subsubsection{Project Structure}

The roadmap comprised 15~phases (PH-19 to PH-33) and 50~tasks distributed across 7~components: web UI~(16~tasks), API~(11), database~(7), testing~(7), configuration~(4), observability~(3), and documentation/release~(2). Tasks were organized into a dependency graph with a defined critical path of 9~tasks, from data persistence to final release.

\subsubsection{Multi-Agent Orchestration}

Four heterogeneous LLM agents were assigned to tasks based on specialization:

\begin{itemize}[nosep]
    \item \textbf{Claude Sonnet 4.6} --- 10~tasks: database schema specification, API contract definition, configuration system design;
    \item \textbf{Codex GPT-5} --- 10~tasks: UI architecture specification, configuration audit, API implementation, testing;
    \item \textbf{Antigravity (Gemini 3 Pro)} --- 5~tasks: persistence implementation, repository layer, service integration;
    \item \textbf{Claude Opus 4.6} --- 5~tasks: security and privacy documentation, observability strategy, deployment guides.
\end{itemize}

\subsubsection{Event Store Analysis}

The execution generated 86~events over approximately 15~hours (2026-02-19 09:00 to 2026-02-20 00:14, \texttt{America/Sao\_Paulo}), with the following distribution: 30~claims, 30~completions, 17~promotions (backlog $\to$ ready), 8~phase completions, and 1~version initialization. All 30~task completions had \texttt{acceptance\_results} with positive validation.

A key observation is the evidence of \emph{concurrent multi-agent execution}: at timestamp \texttt{2026-02-19T21:55}, six claims were recorded within the same minute—Antigravity completing task T-2601 while Claude Opus~4.6 simultaneously claimed five tasks (T-2301, T-2302, T-2303, T-2401, T-2403). This demonstrates that the event store's append-only semantics naturally serializes concurrent agent activities while preserving the temporal ordering necessary for replay.

\subsubsection{Progression and Current State}

At the time of analysis, 31~of~50 tasks (62\%) were completed (\texttt{done}), 2~were in the \texttt{ready} state, and 17~remained in the \texttt{backlog}. Eight out of fifteen phases were completed. The remaining phases cover UI implementation (PH-27 to PH-32) and final hardening/release (PH-33), demonstrating that ESAA can manage an ongoing project with partial completion while maintaining full traceability of concluded work.

\subsection{Comparative Summary}

\Cref{tab:case_comparison} presents a quantitative comparison of the two case studies.

\begin{table}[ht]
\centering
\caption{Quantitative comparison of the two case studies.}
\label{tab:case_comparison}
\small
\begin{tabularx}{\textwidth}{lXX}
\toprule
\textbf{Metric} & \textbf{CS1: Landing Page} & \textbf{CS2: Clinic ASR} \\
\midrule
Total tasks                  & 9              & 50 \\
Total events                 & 49             & 86 \\
Distinct agents              & 3 (composition) & 4 (concurrent) \\
Phases                       & 1 pipeline     & 15 (8 completed) \\
Components                   & 3 (spec/impl/QA) & 7 (DB, API, UI, tests, config, obs, docs) \\
Duration                     & Single session  & $\sim$15 hours \\
\texttt{output.rejected}     & 0              & 0 \\
\texttt{verify\_status}      & ok             & ok (partial---31/50) \\
Concurrent claims            & No             & Yes (6 in 1 min) \\
Dependency graph             & Linear         & DAG with critical path \\
Event vocabulary             & 15 types       & 5 types (simplified) \\
Acceptance criteria          & Implicit       & Explicit per task \\
\bottomrule
\end{tabularx}
\end{table}

The evolution from CS1 to CS2 reveals a natural maturation of the protocol: the event vocabulary was simplified from 15~to~5~action types, suggesting that the core lifecycle (\texttt{claim} $\to$ \texttt{complete} with \texttt{promote} for backlog management) captures essential state transitions. Internal orchestrator events (e.g., \texttt{orchestrator.dispatch}, \texttt{orchestrator.file.write}) can be inferred from the sequence rather than explicitly logged, reducing log verbosity while preserving traceability.

\section{Results and Discussion}
\label{sec:results}

\subsection{Structural Compliance and Reduction of Parsing Failures}

By restricting agent output to a validatable JSON envelope and a small vocabulary of actions, ESAA reduces the probability of parsing failures and "out-of-contract outputs." In both case studies, the \texttt{output.rejected} count was zero, suggesting that the constrained output regime was sufficient to keep the agents (operated by four distinct LLM providers) within the protocol. This strategy is consistent with the goals of structured generation libraries, which prioritize syntactic compliance to enable deterministic execution~\cite{outlines,xgrammar}.

\subsection{Auditability and Debugging through Replay}

The primary output of ESAA is not just the generated code, but the complete trail of events and their effects. This allows for "time-travel debugging" via replay: reprocessing the event store from event zero to reconstruct the read-model exactly and verify divergences. Using SHA-256 hashing as a state projection commitment follows best practices for change detection and message integrity~\cite{nist2015shs}. To reduce serialization ambiguities, canonicalization through key sorting and string normalization aligns with the JSON Canonicalization Scheme~\cite{rfc8785}.

\subsection{Multi-Agent Coordination via Event Store}

The \texttt{clinic-asr} case study demonstrates that the event store serves as a natural coordination mechanism for heterogeneous agents. Three properties emerge from the data:

\begin{enumerate}[nosep]
    \item \textbf{Serialized accountability}: despite concurrent execution, the append-only log preserves total ordering, allowing precise attribution of each state change to a specific agent and task.
    \item \textbf{Specialization tracking}: the log reveals agent specialization patterns—Claude Sonnet handled specification tasks, Codex GPT-5 handled UI and testing, Antigravity handled implementation, and Claude Opus handled cross-cutting concerns. This information is forensically recoverable solely from the event store.
    \item \textbf{Phase-gated progression}: the \texttt{promote} $\to$ \texttt{claim} $\to$ \texttt{complete} $\to$ \texttt{phase.complete} sequence enforces dependency ordering without requiring agents to be aware of each other, reducing coordination overhead.
\end{enumerate}

\subsection{Context Efficiency and Degradation Mitigation}

ESAA proposes that the agent should not carry long-term memory in the raw prompt; it should receive a purified view (roadmap + relevant facts) derived from the log. This approach addresses limitations like \emph{lost-in-the-middle}, as the orchestrator selectively injects information needed for the current step, rather than relying on the model to maintain all details across long windows~\cite{liu2024lost}. In the \texttt{clinic-asr} case, with 50~tasks and complex interdependencies, the purified view mechanism was essential for maintaining agent coherence over the 15~hours of execution.

\subsection{Security and Blast Radius}

By denying direct writing and imposing boundaries per task type, ESAA reduces the blast radius of a compromised agent (e.g., via prompt injection). The strategy aligns with broad recommendations of least privilege and prevention of integrity and logging failures in applications, as synthesized in critical risk guides~\cite{owasp2021}. It is also worth noting that parsing/grammar layers can have availability vulnerabilities when used in high-throughput pipelines; therefore, the implementation must treat grammar engines as components to be monitored and versioned~\cite{gitlab2025xgrammar}.

\subsection{Overhead Considerations}

The ESAA architecture introduces overhead in three dimensions: (i)~\emph{token overhead} from the JSON envelope and validation preamble in each agent invocation (estimated at 200--500 tokens per invocation); (ii)~\emph{latency overhead} from schema validation and event persistence (sub-second per event in both case studies); and (iii)~\emph{storage overhead} of the append-only log (the 86-event log of \texttt{clinic-asr} occupies approximately 15\,KB). These costs are negligible compared to LLM inference costs and provide substantial returns in auditability and reproducibility.

\section{Threats to Validity}
\label{sec:threats}

\textbf{Internal validity.} Both case studies were executed in controlled environments with defined project scopes. Although the \texttt{clinic-asr} pipeline demonstrates significantly higher complexity than the landing page (50~vs.~9~tasks, 4~agents, 7~components), strong causality regarding performance in enterprise repositories cannot be concluded without a larger sample. Furthermore, the use of temperature~0.0 reduces variability but may not reflect setups that explore sampling diversity~\cite{yao2022react}.

\textbf{External validity.} Results from a landing page and a clinical POC do not necessarily generalize to domains with continuous integration pipelines, database migrations, or monorepos at scale. Benchmarks on real issues suggest that repository complexity directly influences agent success, and ESAA must be tested on workloads aligned with SWE-bench and harder variants~\cite{jimenez2024swe, openai2024swe}.

\textbf{Construct validity.} Measures such as \texttt{output.rejected} and \texttt{verify\_status} capture compliance and state reproducibility but do not directly measure "design quality" or "business value" of the software. To broaden constructs, it is recommended to include human evaluations and quality metrics (e.g., test coverage, accessibility, maintainability).

\textbf{Conclusion validity.} Two case studies (n=2) improve upon the initial single-case design but still limit statistical inferences. The appropriate conclusion is that ESAA demonstrates technical feasibility and produces verifiable invariants in scenarios of different complexities, not that it maximizes performance on any benchmark.

\section{Conclusion and Future Work}
\label{sec:conclusion}

ESAA proposes a paradigm shift: treating LLMs as intention emitters under contract, rather than "developers" with unrestricted permissions. By adopting Event Sourcing as the source of truth and applying verifiable projections through replay, the architecture offers native auditability, operational immutability, and state reproducibility~\cite{fowler2005event}. In parallel, the use of formal contracts and constrained output reduces the structure gap and enables a deterministic execution pipeline~\cite{jsonschema2020,outlines}.
The two case studies provide complementary evidence: the landing page demonstrates the architecture's core mechanics in a controlled single-pipeline scenario, while \texttt{clinic-asr} validates multi-agent concurrent orchestration with heterogeneous LLMs, complex dependency graphs, and phase-gated progression in a real-world application domain.

Future work includes: (i)~an official CLI (\texttt{esaa init/run/verify}) with integration to remote repositories; (ii)~conflict detection and resolution strategies for concurrent file modifications; (iii)~automated time-travel debugging with visual diff comparison of projections at arbitrary event points; (iv)~systematic evaluation on real-world issue benchmarks (SWE-bench) and enterprise-scale projects~\cite{jimenez2024swe, openai2024swe}; and (v)~formal verification of orchestrator invariants using model checking techniques.

\bibliographystyle{plain}

\appendix

\section{Agent Output Contract JSON Schema}
\label{app:agent_schema}

\begin{lstlisting}[basicstyle=\ttfamily\scriptsize]
{
  "$schema": "https://json-schema.org/draft/2020-12/schema",
  "$id": "https://esaa.dev/schemas/agent-output-v0.3.0.json",
  "title": "ESAA - Agent Output Contract",
  "type": "object",
  "additionalProperties": false,
  "required": ["schema_version","correlation_id","task_id",
    "attempt_id","actor","action","idempotency_key","payload"],
  "properties": {
    "schema_version": { "const": "0.3.0" },
    "correlation_id": { "type": "string", "minLength": 8 },
    "task_id":        { "type": "string", "minLength": 3 },
    "attempt_id":     { "type": "string", "minLength": 8 },
    "actor":          { "type": "string", "pattern": "^agent-.*" },
    "action":         { "enum": ["agent.result", "issue.report"] },
    "idempotency_key":{ "type": "string" },
    "payload": {
      "type": "object",
      "additionalProperties": false,
      "properties": {
        "summary":   { "type": "string", "maxLength": 2000 },
        "proposals": {
          "type": "array",
          "items": {
            "type": "object",
            "required": ["type", "path", "patch"],
            "properties": {
              "type":  { "const": "file_patch" },
              "path":  { "type": "string",
                         "pattern": "^(\\.roadmap/|src/)" },
              "patch": { "type": "string", "minLength": 1 }
            }
          }
        },
        "issue": {
          "type": "object",
          "required": ["title", "details", "severity"],
          "properties": {
            "title":    { "type": "string" },
            "details":  { "type": "string" },
            "severity": { "enum": ["low","medium","high","critical"] }
          }
        }
      }
    }
  }
}
\end{lstlisting}

\section{Roadmap Read-Model JSON Schema}
\label{app:roadmap_schema}

\begin{lstlisting}[basicstyle=\ttfamily\scriptsize]
{
  "$schema": "https://json-schema.org/draft/2020-12/schema",
  "$id": "https://esaa.dev/schemas/roadmap-v0.3.0.json",
  "title": "ESAA - Roadmap Read-Model",
  "type": "object",
  "required": ["schema_version","project","run","tasks","indexes"],
  "properties": {
    "schema_version": { "const": "0.3.0" },
    "project": {
      "type": "object",
      "required": ["name", "created_at", "audit_scope"],
      "properties": {
        "name":        { "type": "string" },
        "created_at":  { "type": "string" },
        "audit_scope": { "type": "string" }
      }
    },
    "run": {
      "type": "object",
      "required": ["run_id","status","last_event_seq",
                    "last_event_ts","verify_status"],
      "properties": {
        "run_id":       { "type": "string" },
        "status":       { "enum": ["initialized","running",
                                    "success","failed"] },
        "last_event_seq":           { "type": "integer", "minimum": 0 },
        "last_event_ts":            { "type": "string" },
        "projection_hash_sha256":   { "type": "string",
                                      "pattern": "^[a-f0-9]{64}$" },
        "verify_status":            { "enum": ["ok","mismatch",
                                                "corrupted"] }
      }
    },
    "tasks": {
      "type": "array",
      "items": {
        "type": "object",
        "required": ["task_id","kind","title","state",
                     "depends_on","files"],
        "properties": {
          "task_id":    { "type": "string" },
          "kind":       { "enum": ["spec","impl","qa",
                                    "emergency_patch"] },
          "title":      { "type": "string" },
          "state":      { "enum": ["todo","in_progress",
                                    "blocked","done"] },
          "depends_on": { "type": "array",
                          "items": { "type": "string" } },
          "files":      { "type": "array",
                          "items": { "type": "string" } }
        }
      }
    },
    "indexes": { "type": "object" }
  }
}
\end{lstlisting}

\section{Canonical Event Vocabulary}
\label{app:event_vocab}

\Cref{tab:event_vocab_full} presents the complete event vocabulary (v0.3.0) as specified in the architecture, and \Cref{tab:event_vocab_simplified} presents the simplified vocabulary observed in the \texttt{clinic-asr} case study.

\begin{table}[ht]
\centering
\caption{Complete canonical event vocabulary (v0.3.0).}
\label{tab:event_vocab_full}
\small
\begin{tabularx}{\textwidth}{lX}
\toprule
\textbf{Event} & \textbf{Description} \\
\midrule
\texttt{run.init}                & Initializes a run (metadata and scope). \\
\texttt{attempt.create}          & Opens an attempt for a task and links an agent. \\
\texttt{attempt.timeout}         & Expires an attempt (TTL) without completing the task. \\
\texttt{orchestrator.dispatch}   & Triggers agent execution for an attempt. \\
\texttt{agent.result}            & Agent intention (patch proposals), validated by schema/boundary. \\
\texttt{issue.report}            & Record of defect/incident; may generate a hotfix task. \\
\texttt{output.rejected}         & Rejected output (violated schema, boundary, authority, or idempotency). \\
\texttt{orchestrator.file.write} & Applied effect: writing to authorized files. \\
\texttt{orchestrator.view.mutate}& Applied effect: read-model update (derived). \\
\texttt{task.create}             & Creates a new task (includes hotfix). \\
\texttt{task.update}             & Updates task state (e.g., todo $\to$ in\_progress $\to$ done). \\
\texttt{verify.start}            & Starts audit via replay + deterministic hashing. \\
\texttt{verify.ok}               & Audit passed; records \texttt{projection\_hash\_sha256}. \\
\texttt{verify.fail}             & Audit failed; records divergence (expected vs computed). \\
\texttt{run.end}                 & Finalizes run (success/failed) and closure criteria. \\
\bottomrule
\end{tabularx}
\end{table}

\begin{table}[ht]
\centering
\caption{Simplified event vocabulary observed in \texttt{clinic-asr} (CS2).}
\label{tab:event_vocab_simplified}
\small
\begin{tabularx}{\textwidth}{lXr}
\toprule
\textbf{Event} & \textbf{Description} & \textbf{Count} \\
\midrule
\texttt{roadmap.version} & Initializes the roadmap version. & 1 \\
\texttt{promote}         & Moves a task from backlog to ready. & 17 \\
\texttt{claim}           & Agent claims a task for execution. & 30 \\
\texttt{complete}        & Agent completes a task with acceptance results. & 30 \\
\texttt{phase.complete}  & All tasks in a phase are finished. & 8 \\
\bottomrule
\end{tabularx}
\end{table}

The simplification from 15 to 5 event types reflects a design insight: internal orchestrator events (\texttt{dispatch}, \texttt{file.write}, \texttt{view.mutate}) can be inferred from the \texttt{claim} $\to$ \texttt{complete} sequence, reducing log verbosity while maintaining essential traceability guarantees. The full vocabulary remains available for environments requiring maximum granularity.

\section{Canonicalization and Verify Pseudocode}
\label{app:pseudocode}

\begin{lstlisting}[language=Python,basicstyle=\ttfamily\scriptsize]
import json
import hashlib

def canonical_json(obj):
    # Deterministic JSON: sort keys, no spaces, UTF-8, trailing LF
    return (
        json.dumps(
            obj,
            sort_keys=True,
            separators=(",", ":"),
            ensure_ascii=False,
        ).encode("utf-8")
        + b"\n"
    )

def hash_input(projected_state):
    # Avoid self-reference: exclude run metadata and hash field
    return {
        "schema_version": projected_state["schema_version"],
        "project": projected_state.get("project"),
        "tasks": projected_state["tasks"],
        "indexes": projected_state.get("indexes", {}),
    }

def compute_projection_hash(projected_state):
    return hashlib.sha256(
        canonical_json(hash_input(projected_state))
    ).hexdigest()

def esaa_verify(events, roadmap_json):
    projected = project_events(events)  # pure function
    computed = compute_projection_hash(projected)
    stored = roadmap_json["run"]["projection_hash_sha256"]
    return (
        {"verify_status": "ok"}
        if computed == stored
        else {"verify_status": "mismatch"}
    )
\end{lstlisting}

\section{Representative Extract from \texttt{activity.jsonl} (CS2: clinic-asr)}
\label{app:event_extract}

The following extract illustrates concurrent multi-agent execution in the \texttt{clinic-asr} event store (lines 58--68), showing Claude Opus~4.6 claiming five tasks simultaneously:

\begin{lstlisting}[basicstyle=\ttfamily\scriptsize,breaklines=true]
{"ts":"2026-02-19T21:55:44","action":"claim","task_id":"T-2301",
 "agent_id":"claude-opus-4-6","acceptance_results":null}
{"ts":"2026-02-19T21:55:45","action":"claim","task_id":"T-2302",
 "agent_id":"claude-opus-4-6","acceptance_results":null}
{"ts":"2026-02-19T21:55:46","action":"claim","task_id":"T-2303",
 "agent_id":"claude-opus-4-6","acceptance_results":null}
{"ts":"2026-02-19T21:55:47","action":"claim","task_id":"T-2401",
 "agent_id":"claude-opus-4-6","acceptance_results":null}
{"ts":"2026-02-19T21:55:48","action":"claim","task_id":"T-2403",
 "agent_id":"claude-opus-4-6","acceptance_results":null}
{"ts":"2026-02-19T21:56:10","action":"complete","task_id":"T-2301",
 "agent_id":"claude-opus-4-6",
 "acceptance_results":{"Rules documented":true}}
{"ts":"2026-02-19T21:56:20","action":"complete","task_id":"T-2302",
 "agent_id":"claude-opus-4-6",
 "acceptance_results":{"Events documented":true}}
\end{lstlisting}

\end{document}